\documentclass[10pt, oneside]{article}
\usepackage[top= 3.8cm, bottom=3.8cm, left=2.6cm, right=2.6cm]{geometry}
\usepackage{graphicx}
\usepackage{setspace}
\usepackage{titlesec}
\usepackage{wrapfig}
\usepackage{subfig}
\usepackage{url}
\usepackage{natbib}
\usepackage{color}

\begin{document}
\begin{titlepage}

\title{The Cyborg Astrobiologist: Porting from a wearable computer
to the Astrobiology Phone-cam}

\author{Alexandra Bartolo \\
\footnote{corresponding author: {\it Email:} abbart@eng.um.edu.mt
\hspace{12cm}\break{\it Telephone:} +356 2340 2570
{\it Fax:} +356 21 343577\hspace{12cm}\break
\vspace{0.1cm}\break
Publication information: Received xxxx 2006, xxxx,\hspace{10cm}\break
by the {\it International Journal of Astrobiology}, xxxxx
 }
{\it Department of Electronic Systems Engineering}\\
{\it University of Malta, Malta}\\
\and
Patrick C. McGuire\\
{\it (currently at) McDonnell Center for the Space Sciences}\\
{\it Department of Earth and Planetary Sciences and Department of Physics}\\
{\it Campus Box 1169, Washington University, St. Louis, 63130 USA}\\
{\it (formerly at) Robotics \& Planetary Exploration Laboratory; Transdisciplinary Laboratory}\\
{\it Centro de Astrobiolog\'ia (INTA/CSIC)}\\
{\it Instituto Nacional T\'ecnica Aeroespacial}\\
{\it Carretera de Torrej\'on a Ajalvir km 4.5, Torrej\'on de Ardoz, Madrid, Spain 28850}\\
\and
Kenneth P. Camilleri, Christopher Spiteri, Jonathan C. Borg, Philip J. Farrugia\\
{\it Department of Electronic Systems Engineering}\\
{\it University of Malta, Malta}\\
\and
Jens Orm\"o\\
{\it Planetary Geology Laboratory}\\
{\it Centro de Astrobiolog\'ia (INTA/CSIC), Instituto Nacional T\'ecnica Aeroespacial}\\
{\it Carretera de Torrej\'on a Ajalvir km 4.5, Torrej\'on de Ardoz, Madrid, Spain 28850}\\
\and
Javier G\'omez-Elvira, Jos\'e Antonio Rodriguez-Manfredi\\
{\it Robotics \& Planetary Exploration Laboratory}\\
{\it Centro de Astrobiolog\'ia (INTA/CSIC), Instituto Nacional T\'ecnica Aeroespacial}\\
{\it Carretera de Torrej\'on a Ajalvir km 4.5, Torrej\'on de Ardoz, Madrid, Spain 28850}\\
\and
Enrique D\'iaz-Mart\'inez\\
{\it (currently at) Direcci\'on de Geolog\'ia y Geof\'isica}\\
{\it Instituto Geol\'ogico y Minero de Espa\~na}\\
{\it Calera 1, Tres Cantos, Madrid, Spain 28760} \\
{\it (formerly at) Planetary Geology Laboratory, Centro de Astrobiolog\'ia}\\
\and
Helge Ritter, Robert Haschke, Markus Oesker, J\"org Ontrup\\
{\it Neuroinformatics Group, Computer Science Department}\\
{\it Technische Fakult\"at}\\
{\it University of Bielefeld}\\
{\it P.O.-Box 10 01 31}\\
{\it Bielefeld, Germany 33501}\\
}

\date{\today}
\vspace{3.0cm}

\maketitle
\thispagestyle{empty}
\end{titlepage}
\pagenumbering{arabic}
\begin{abstract}
We have used a simple camera phone to significantly improve
an `exploration system' for astrobiology and geology. This camera phone
will make it much easier to develop and test computer-vision algorithms
for future planetary exploration. We envision that the `Astrobiology Phone-cam'
exploration system can be fruitfully used in other problem domains as well.
\end{abstract}

\thispagestyle{myheadings}
\markright{The Cyborg Astrobiologist: Porting to the Astrobiology Phone-cam: Bartolo et al.}
\newpage

\section{Introduction}
\begin{spacing}{1}

Planetary exploration by autonomous robotic systems cannot be carried out successfully unless significant testing of the underlying computer vision algorithms is performed. In our previous work, we have demonstrated the use of a wearable computer system, the \emph{Cyborg Astrobiologist}, capable of testing computer-vision algorithms as part of semi-autonomous exploration systems at remote geological and astrobiological field sites~\citep{mcguire_IJA04,mcguire_IJA05}. In that work, we showed that the exploration system, which was based upon newly-developed `uncommon maps' and previously-developed `interest maps'~\citep{rae02,mcguire02}, could viably and robustly be utilized during remote field missions to localize interesting geochemical or hydrological features. Our system carries out the navigation process using the lower end of the spectral resolution, making use of three colour imagery to distinguish between regions of unusual colour. Navigation using higher spectral resolution spectrometry, for example, navigation based on mineralogical differences, will yield more interesting results but this is beyond the scope of the current work.

\begin{figure}
\begin{centering}
\subfloat[\emph{Wearable computer and camera.}] {\label{fig:CameraPhoneCapture}\includegraphics[scale = 0.47]{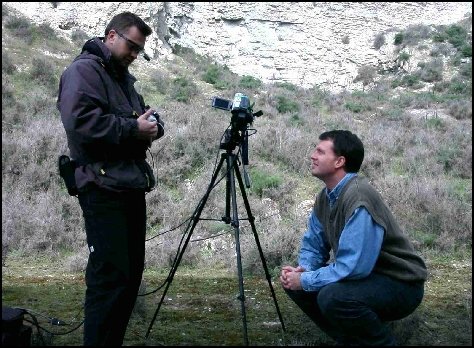}}
\subfloat[\emph{Camera phone.}] {\label{fig:CameraPhoneView}\hspace{0.4cm}\includegraphics[scale = 0.475]{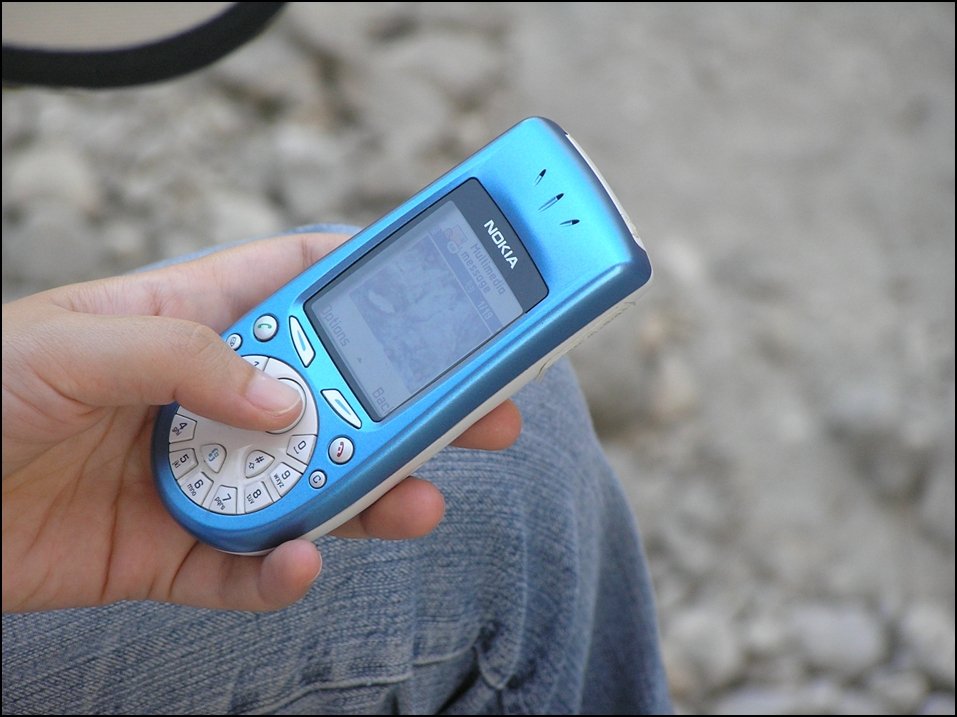}}
\caption{\emph{Comparison of the equipment required when using a wearable computer to that required when using a camera phone.}}\label{CameraPhone}
\end{centering}
\end{figure}

In this work, we report upon the development and initial field tests of one of the recent enhancements of the Cyborg Astrobiologist system, namely its porting from the wearable computer and connected video camera into a remote server and a camera phone (mobile phone with an inbuilt digital camera). By using a camera phone instead of a wearable computer, we offer the user several advantages including: considerable reduction in the equipment required during exploration (Figure~\ref{CameraPhone}), less special training to use the system, and access to higher-speed computational servers. However, the inbuilt cameras in mobile phones are small devices intended to provide basic image capturing facility. For this reason, camera phones do not allow the use of peripheral computer-controlled devices, such as a robotic pan-tilt mount, a zoom lens, or a digital microscope, posing a limit on the imaging capabilities. The camera-phone will also incur a modest increase in the time elapsed before the computer-vision results may be viewed. This is due to the fact that the phone cannot perform the necessary image processing and images must be transmitted to a remote server.

Although the quality of the cheaper camera-phone models cannot be compared with that of professional digital cameras, improvements in mobile-phone technologies are introducing significant improvements in the imaging capabilities of camera phones. In fact, the most recent camera phone models, such as Nokia's E90, boast of a $3.2$~Megapixel resolution with flash and autofocus features. This means that the image quality of camera phones may soon be comparable with that of digital cameras. The considerable reduction in the equipment required during field testing makes it easier for users to complete a remote field testing of the computer-vision exploration system.

Within this problem domain, the fruits of this project could be adapted to serve as the computer vision system for the microbot swarm for planetary surface and subsurface exploration, as envisioned by \citet{Dubowsky}, or the lowest tier sensor-web of a multi-tier exploration system, as envisioned by \citet{Fink}. Furthermore, such a camera-phone system could readily be adapted for a number of other applications or exploration algorithms beyond the application and algorithm which we have developed and tested.

This paper is structured as follows: Section 2 describes the integration of the camera phone with the \emph{Cyborg Astrobiologist} system, Section 3 gives a description of the image processing involved, wherein we summarize our past work, whilst Section 4 gives the results obtained during initial field tests. Possible future work is presented in Section 5, finalizing the paper with Section 6 which presents the conclusions derived from this work.

\section{Camera Phone Interface}
Mobile technology has been developed over the years such that today even the cheapest mobile phones have an inbuilt digital camera and simple web-browsing capabilities. Advances in mobile communication also allow mobile users to exchange pictures using multimedia messaging service (MMS), which was originally intended as a fun and interesting alternative to normal communication. Camera-phones and MMS have been used to increase the communication between designers giving the possibility for designers to have remote access to software that interprets the designer's drawing \citep{philip_SBM04}. A similar approach may be used for field exploration, allowing the user to explore a potential geological interesting site using just a camera-phone.

\begin{figure}[t!]
\begin{centering}
\includegraphics[scale = 0.7]{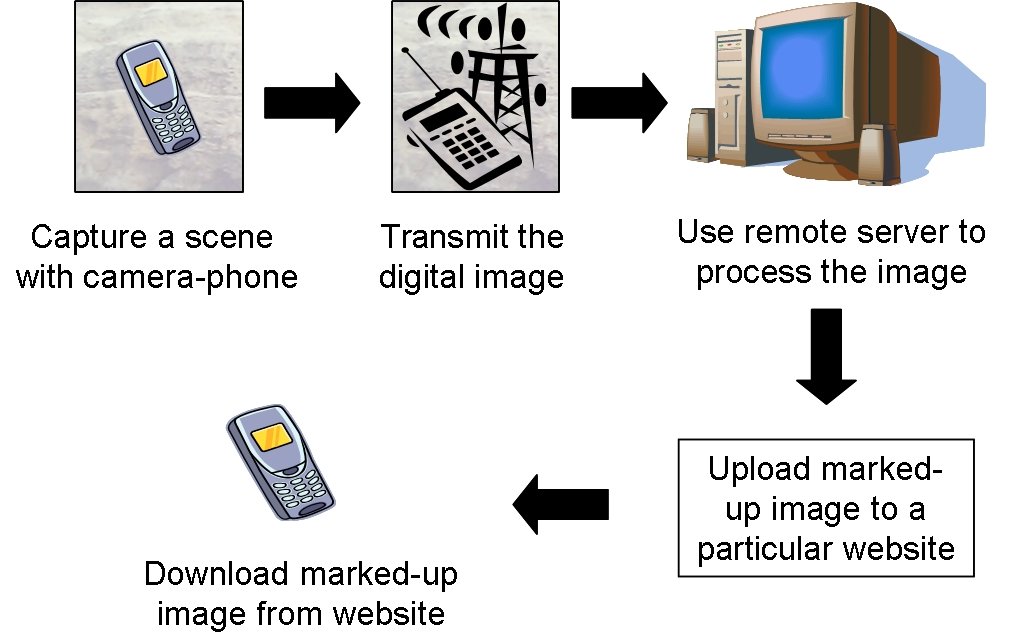}
\caption{\emph{The framework that enables the use of camera-phones for remote field exploration.}}\label{framework}
\end{centering}
\end{figure}

As shown in Figure~\ref{framework}, the field explorer takes an image of a geological or astrobiological scene using the camera-phone. This picture is then sent to a particular e-mail address, as a mail attachment, using MMS. A remote server is used to automatically and periodically check for incoming mail. A new e-mail will initiate the computer-vision system which searches for uncommon interest points. These interest points are reported back to the field explorer in the form of a marked-up image that is made available on a particular web-page. The field explorer may download and view this marked-up image on the camera-phone. Since communication between the camera phone and the remote computer takes place via multimedia messaging, no additional hardware is required to act as an interface between the camera phone and the computer acting as a remote server.

This framework requires that the remote server has installed an automated mail-watcher in order to periodically check for incoming mail. This has been implemented by using Microsoft Outlook$^{\textregistered}$ and Microsoft Visual Studio$^{\textregistered}$. Microsoft Outlook offers a \emph{scheduled send/receive} option which automatically and periodically checks for and downloads incoming mail. The actual mail-watcher software has been implemented through Microsoft Visual Studio in a similar manner as that used in the InPro system~\citep{Inpro}. Using Microsoft Visual Studio, it is possible to access and process all e-mails that Microsoft Outlook downloaded onto the remote server. For security purposes, the mail-watcher filters all e-mails, retaining those containing an attachment whose name starts with a specific prefix string. This prefix string is entered by the field explorer when saving the image on the camera phone, prior to transmitting it as an MMS. When such an e-mail is detected, the attached image is automatically saved onto the remote server's hard-drive from where it may be accessed by the image-processing software.

In our implementation, the NEO software~\citep{NEO} is used to carry out the required image-processing tasks. The NEO software is configured such that it automatically shuts down after all processing is completed, returning the CPU control to the mail-watcher. This has the additional task of loading the marked-up image onto a specific web-page before checking the mail inbox for newly saved mail. In this way, the exploration process may be monitored by the geologist on site, and any other geologist interested in the exploration process, even if they are not on site. This can be particularly useful for international teams working through a network without being physically in the same city or continent.

Connectivity between the camera-phone and the remote server through MMS has been found suitable for regions that have good mobile network coverage. However, there are field sites, such as caves, remote mountaintops, or the cold deserts of Antarctica, where insufficient network coverage would limit the communication between the camera-phone and the server. This may be amended by using other forms of wireless communication, which would however require that the remote server is also located on site. For example, most modern mobile phones offer the possibility of communicating via bluetooth technology. In this case, the image transmitted by the mobile phone is saved directly onto the server's hard drive. This implies that the mail-watcher, rather than monitoring Microsoft Outlook's inbox, will be used to monitor a folder on the server's hard drive. Although this form of communication makes the system less mobile than communication via MMS, the system will still give the field explorer greater mobility than the wearable computer system described in \citet{mcguire_IJA05}.

In either communication mode, the system can monitor, process and transmit images without the need of human intervention. This opens up the possibilities of remote and automated navigation, particularly since marked-up images can be made available to other persons apart from the field explorer.

\section{Computer Vision with Uncommon Maps}

The computer vision software that uses uncommon maps and interest maps is an extension of the software used by the GRAVIS robot in Bielefeld, Germany~\citep{rae02,mcguire02}. Many of the extensions were made as part of the Cyborg Astrobiologist project\footnote{Refer to http://www.cab.inta.es/$\sim$CYBORG/cyborg.summary.html for more details on the Cyborg Astrobiologist project.}. The software for the GRAVIS robot focused upon the three-dimensional detection of pointing-finger gestures and toy blocks for human-machine cooperation research in a controlled indoor environment. The challenge was to determine the interest points for the active vision system of the robot in a dynamic environment, which often included verbal and gestural requests from the human to the robot. General capabilities for image segmentation were not required. Likewise, general capabilities for finding the uncommon points of the images were not necessary for the GRAVIS robot. However, somewhat general capabilities of finding interesting points of the images were essential for the GRAVIS system. The interest map was implemented by summing 6-8 different maps in a dynamic way. The 6-8 different maps in the interest-map sum were each salient features in themselves, such as skin-colour, motion, edges, or colour saturation. The resultant interest map was a rather robust way for the GRAVIS robot to find interesting areas of the image for the controlled, artificial environment in its domain.

With this knowledge base, we decided to extend and adapt the GRAVIS interest-map technique to include the processing of new uncontrolled, natural environments. Such environments would be the domain of planetary rovers or borehole-inspection systems, thus allowing the GRAVIS system to find interesting targets on or underneath a planetary surface. We decided that:
\begin{itemize}
\item the platform for testing the system would be a wearable computer connected to a digital video camera;
\item one of the main areas of software development should be in image segmentation -- this is essential for capturing part of the visual thought processes of practicing human geologists;
\item as a first step in the Cyborg Astrobiologist research program, we would develop a computer vision system that would be capable of detecting the uncommon areas of the images -- often in geological outcrops, human geologists are most drawn towards those parts of the outcrop which are most different from the remainder of the exposed rocks, as it often is the relation between the anomalous parts and the common background that reveals the geological history of the outcrop (e.g. a magmatic dike cutting an older rock).
\end{itemize}

These basic decisions for the directions of the Cyborg Astrobiologist research program led to a computer-vision system with a 3-layer interest map, similar to the GRAVIS architecture, but with each layer of the interest map being an `uncommon map'. The uncommon map was based upon looking for small areas in a segmentation of the image. Three different image segmentations, one for hue, one for saturation and one for intensity, provided the inputs to the uncommon-map algorithm. Remarkably, this simple computer-vision algorithm more than often found interesting points which agreed with the interest points found by humans or even human geologists~\citep{mcguire_IJA05}. We attribute this robustness and agreement with human judgement to be due to the simplicity of the algorithm and perhaps also due to a rough correspondence between the computer-vision algorithm and some of the low-level visual processes of humans.

\subsection{Image Preprocessing Required}
A slight discrepancy exists between the size of the images taken by the mobile phone and the standard processing size of our computer vision system. In order to limit the computer processing time to under two minutes, which is the specification given by the patience of a typical user, the computer vision system uses a standard image size of $192 \times 144$ pixels. However, mobile phone images have a size of $640 \times 480$ pixels. Thus, the images were automatically cropped to $576 \times 432$ and then downsampled by 3 in both directions, in order to fit into the $192 \times 144$ standard processing size. As will be shown in the next section, this three-fold reduction of image resolution is tolerable, and has not yet had a significant impact upon our field tests.

\subsection{Modular Graphical Programming}
The computer vision software was programmed in the NEO Graphical Programming Language, which was developed in Bielefeld, Germany~\citep{NEO}. We are using the version of NEO that works in Microsoft Windows$^{\textregistered}$. The modularity offered by NEO and the encapsulation tools within NEO facilitates the programming required for complex tasks like the image segmentation and uncommon maps that we have implemented here. Furthermore, the modularity and ease-of-adaptation have made NEO a key tool for this project, allowing us to easily adapt code from other NEO projects.

\section{Analysis of System Performance}
\begin{figure}
\begin{center}
\subfloat[\emph{The red box indicates the region focused on during exploration. To capture Image A of the image sequence shown in Figure~\ref{mission2}, the operator of the Astrobiology Phone-cam was standing roughly at the place marked with the cross.\label{global}}]{\includegraphics[scale = 0.4]{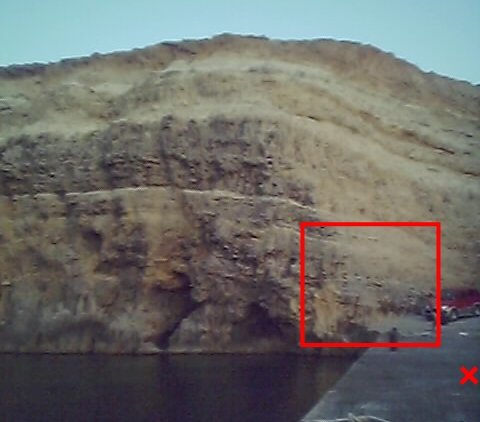}}\\
\subfloat[\emph{Close up of the region studied. The rock region has been partially segmented to show the different colours in the rocks. Areas enclosed in blue indicate the presence of microbiotic crusts, those in green show cavities, magenta indicated lighter calcite regions and red indicates oxidation effects. \label{segments}}]{\includegraphics[scale = 0.3] {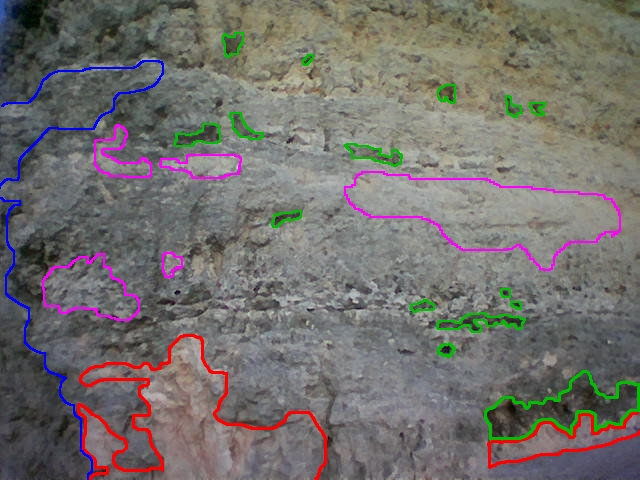}}\\
\caption{\emph{Field Mission at Anchor Bay, Malta.}}
\label{GlobalView}
\end{center}
\end{figure}

Figure~\ref{GlobalView} shows a cliff side in Anchor Bay (Malta) where one testing mission with the Astrobiology Phone-cam was conducted. The cliff consists of Upper Coralline Limestone sediments, indicating the remains of ancient reefs which often contain rhodoliths of coralline algae. Figure~\ref{global} shows that the strata are contorted towards the right, due to a nearby subsidence. This is typical for this region of the island which is characterized by horst and graben structures. As shown in Figure~\ref{segments} the rocks in this site show three main colours, namely dark areas which may be due to either microbiotic crusts or cavities, lighter coloured gray areas which are exposures of calcite and reddish regions which are a surface effect due to oxidation of iron. This site contains characteristics similar to those found at sites near Rivas Vaciamadrid and Riba de Santiuste, both in Spain, which have been studied previously with the wearable computer as part of the Cyborg Astrobiologist research project~\citep{mcguire_IJA04,mcguire_IJA05}. Those sites contained mainly grey and white gypsum and clay deposits (Rivas Vaciamadrid) and reddish sandstones (Riba de Santiuste), with occasional colour from water runoff, geochemical oxidation-reduction processes, or microbiotic crusts.

\begin{figure}
\center{\includegraphics[height=18.0cm]{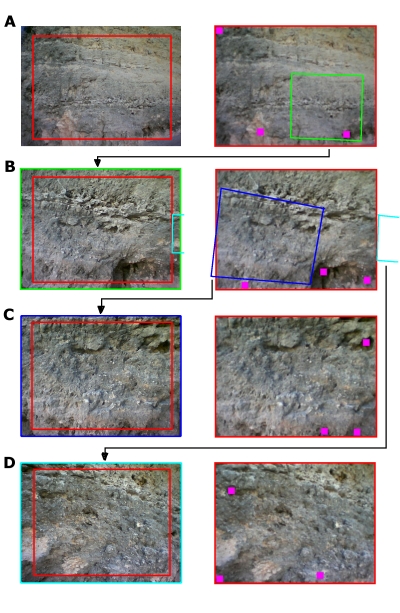}}
\caption{\emph{Sequence of images acquired and analyzed in real-time by the Astrobiology Phone-cam. Each image is captured at a resolution of $96~\emph{dpi}$ and has a size of $640\times480$ pixels. The left column of four images contains the images that were sent by the phone-cam to the remote computer server for analysis. The red boxes in each image on the left are the areas of the images that were analyzed by the uncommon maps in the server's NEO software. The right column of four images contains the images that were sent back to the phone-cam by the remote computer after the uncommon map analysis -- the 3 purple-coloured points were the most interesting points for each image. The processing shown in the right column includes cropping to the red boundary shown in the left image, and subsequent down-sampling by three in each direction, in order to reduce processing time. The green-, blue-, and cyan-coloured boxes represent the extent of the subsequent images. Note that, due to physical constraints upon camera location, the human user was not always able to photograph the area of the image that the computer vision found interesting, but instead she photographed an area of the image that was near the computer-vision interest points.}}
\label{mission2}
\end{figure}

Table~\ref{imagetable} enumerates the images acquired during the testing mission. These images, shown in Figure~\ref{mission2}, show details of smaller beds and laminations in the rock itself. Each bed represents an individual episode of sedimentation or chemical enrichment which in some cases may also result from the selective erosion of pre-existing sediments. The bedding planes are inclined towards the lower right hand side of the each picture, following the general pattern in the rock shown in Figure~\ref{GlobalView}. The small cavities that are visible in the images represent removal of sediment through selective erosion and, as a general impression, seem to be generally correlated with the extent of one particular plane that is probably composed of less indurated material than the layers above and below it.

The left-hand column of Figure~\ref{mission2} contains the images that were sent by the Astrobiology Phone-cam to the remote server computer for processing. These images have a print size of $480 \times 640$ pixels and a resolution of $96 dpi$ and, as explained in the previous section, require down-sampling in order to fit the standard size assumed by the computer vision algorithms. It is interesting to note that these images are largely devoid of substantial flat-field or pixelation effects that would affect our current and foreseen applications. One may also observe that the images do not suffer from image stabilization (jitter) that is normally observed when humans take digital photographs without making use of camera stands. This is mainly due to the fact that the camera-phone model used did not have the capabilities of adjusting the exposure time, taking images instantly. Furthermore, the field tests were carried out in good daylight conditions, such that image stabilization was not an issue even with a more sophisticated digital camera.

The right-hand column of Figure~\ref{mission2} shows the images received by the field operator. These images were received after a delay of about 4-6 minutes which corresponds to the time taken by the remote server to receive the images, process them and make them available to the operator of the Astrobiology Phone-cam. This delay is caused by two factors, namely the delay caused by the mobile service provider to transmit the image and the delay that corresponds to the processing time of the image. This latter delay is not directly related to the phone-cam and is similar to the delay experienced when using the wearable computer. Thus, the porting the Cyborg Astrobiologist system from a wearable computer to a phone-cam system will only increase the time between image capture and viewing of the results by approximately 2-3 minutes.

Using the annotated result images, the human operator then decided how to use the information given by the three interest points from the computer vision, in order to better explore the geological site. In this particular case, due to the physical constraints of the water next to the beach, the human operator could not easily point the Astrobiology Phone-cam to center upon the interest points of Image B, but the operator was able to point the Astrobiology Phone-cam towards areas near those interest points. Note that, unlike the wearable computer system, the human operator has the possibility of sending more than one image to the remote server. The human operator exploited this possibility when transmitting image D which was sent before the result of image C was available. This is beneficial to the human operator who may want to further explore more than one interest point given in a previous result.

From a computer-vision perspective, the system did well. In Image A, the uncommon maps of the remote server's computer-vision system found the localized reddish area in the lower left to be interesting, as well as the darkest two parts of the dark areas, ignoring the bland tan colours to the upper right. Due to physical constraints, the human operator chose to point the Astrobiology Phone-cam at the dark spot chosen by the computer on the lower right of Image A. In the resulting Image B, the remote server's computer-vision system found the dark hole to be interesting, as well as an area to the lower right that had a juxtaposition of reddish colouring and dark colouring. A third point to the lower left was somewhat different than the remainder of the image, so that is why it was chosen; in this case, there was a juxtaposition of brighter white-coloured minerals and a smooth tan-coloured texture.

The human operator of the Astrobiology Phone-cam decided that the dark hole was not interesting, so she tried to explore the other two points in Image B. Unfortunately, due to physical constraints, she was only able to point the camera {\it near} the other two points, instead of centering upon those two points. The resulting images are shown in Image C and Image D. In Image C, the system concentrates on the darker areas of the image -- the upper dark area being a hole and the lower dark areas perhaps being a microbiotic crust. In Image D, the system finds a darker hole in the upper part of the image, a darker microbiotic crust-like area in the lower left of the image, and a bright white crystalline-like area in the lower right of the image.

The image-segmentation software that we use to make the uncommon maps does not yet have texture or colour-texture segmentation capabilities. Despite this fact, the uncommon mapping software did reasonably well at finding the most unique areas in each individual image. This is almost obvious by inspection of Images A, C, D, for at least two of the chosen three points in each image. In Image B, the system also did well, especially after realizing that the uncommon-mapping software rightly ignores the hole-ridden, textured area that dominates an area just above the mid-point of the image. The software rightly ignores this area despite its lack of texture-segmentation capabilities because it is just a juxtaposition of two relatively common colours or shades: bright white and dark gray.

In Image D, one might think that the computer vision would find the large red spot to the lower left of the image to be interesting. The computer-vision system would find such areas interesting if the system were biased to find red spots or large contiguous areas to be interesting. However, our software does not have these biases, focusing instead on identifying those areas that are relatively rare in the image. In this particular image, the reddish colour occurs quite frequently and so it is not chosen by the software.

\section{Future Work}
We intend to further test the system at sites of astrobiological or geological interest. We can learn more about how to optimize the system for the quickest computer-vision processing and for the quickest delivery of MMS messages. We also intend to port a novelty-detection neural network algorithm~\citep{Novelty2,Novelty1} from the wearable computer to the Astrobiology Phone-cam. This novelty-detection neural network was tested on the wearable computer at Rivas Vaciamadrid in the summer of 2005, but we have not yet ported it to the Astrobiology Phone-cam primarily because our implementation of the novelty-detection software needs to be further improved. The improvements include the storage of the novelty-detection neural network memories to the hard disk instead of in volatile RAM memory. The phone-cam NEO software currently is not stored in memory indefinitely; it is reloaded upon receipt of each MMS image. Hence, the storage of the neural network memories to hard disk is a necessary improvement prior to using the novelty-detection software in the astrobiology phone-camera.

Further enhancements in the low-level processing of the images will also be pursued, including real-time calibration of the images for lighting and shading effects \citep{goldman,PiletGLLF06}, as well as enhancements of the image-segmentation algorithm for the segmentation of coloured textures in the images~\citep{freixenet}. With anticipated computer-vision enhancements such as these, we may need to revisit our choice of consumer-grade cameras like the phone-camera discussed in this paper or the digital video camera discussed in previous work. However, there is a great deal of computer-vision development and testing that can be done with the image quality of consumer-grade cameras, so we will tackle this issue at the appropriate moment in the future.

Another enhancement that could be useful in implementing for the uncommon map system  would be to tell the software to ignore those areas found by the uncommon map to be interesting if those types of areas are already found to be interesting in another part of the image. This could be implemented by using a filter to compare certain image features for each of the three uncommon interest point {\it after} those uncommon interest points are determined\footnote{This filter is similar to the previously-described novelty-detection neural-network algorithm, but it filters interest points on a single image, instead of on a sequence of images.}. With this additional filter, the system could bias the user to study truly novel areas in subsequent image acquisition instead of repeatedly studying similar areas.

\section{Conclusion}
The Astrobiology Phone-cam system is a promising platform for testing computer-vision algorithms for planetary exploration. It is miniaturized and more ergonomic than the previous wearable-computer system of the Cyborg Astrobiologist. This facilitates the field exploration and navigation by reducing the burden that the field explorer has to carry. Using the camera phone, Astrobiology Phone-cam system not only has a simpler front-end image capture device, but also an automated image processing procedure which is being carried out on the back-end remote server. In this way the system is made easier to use, requiring no special training or human monitoring. Whilst the computer vision software is essentially the same as that used in our previous systems, the new communication interface between the camera and the image processing software gives the field explorer greater flexibility. Although images are processed sequentially, the field explorer does not need to wait for the results in order to transmit a new image. Thus, the field explorer, may explore multiple points of interest simultaneously. 

We expect that the Astrobiology Phone-cam will allow us to perform field tests more easily, so that we can upgrade the computer vision software in the near future. We intend to use the Astrobiology Phone-cam system instead of the wearable-computer system for much of our future work in the Cyborg Astrobiologist research program.

\section{Acknowledgements}
We would like to acknowledge the support of other research projects which helped in the development of the Astrobiology Phone-cam. Integration of the camera-phone with an automated mail watcher was carried out under the `Innovative Early Stage Design Product Prototyping' (InPro) project, supported by the University of Malta under research grant IED 73-529-2005. Many of the extensions to the GRAVIS interest-map software, programmed in the NEO language, were made as part of the Cyborg Astrobiologist project from 2002-2005 at the Centro de Astrobiologia in Madrid, Spain, with support from INTA and CSIC, and from the Spanish Ramon y Cajal program.

Patrick McGuire acknowledges support from a Robert M. Walker fellowship in Experimental Space Sciences from the McDonnell Center for the Space Sciences at Washington University in St. Louis.

We are grateful for conversations with Peter Halverson and Virginia Souza-Egipsy Sanchez, which were part of the motivation for developing the Astrobiology Phone-cam, and with Sandro Lanfranco who explained the geological features present in Anchor Bay.

\bibliographystyle{apalike}
\bibliography{IJABartolo2}
\newpage

\begin{table}[t!]
\begin{center}
\begin{tabular}{|c|p{2cm}|p{2cm}|p{2cm}|p{2cm}|p{4cm}|}
\hline
\textbf{Image} & \textbf{Distance (m)} & \textbf{Capture Time}& \textbf{Receive Time} & \textbf{Completion Time} & \textbf{Notes by human operator}\\[0.5ex]\hline\hline\
A & 3.0  & 18:58 & 19:00 & 19:02 & The top point is a bit too high and I couldn't get too close to the middle one because of the sea. So I moved towards the one on the right. \\
\hline
B & 1.5 & 19:05 & 19:09 & 19:11 & Middle point is just a hole in the rock. At this distance I can see that there is nothing there. But I took a closer view of the other two points. \\
\hline
C & 0.7 & 19:20 & 19:23 & 19:26 & Looking towards the point indicated on the right in the previous image. \\
\hline

D & 0.7 & 19:22 & 19:25 & 19:28 & Looking towards the point indicated on the left. \\
\hline
\end{tabular}
\caption{\emph{List of images and their attributes for the observing run at Anchor Bay, Malta. The capture time indicates the time at which each image was taken, the receive time gives the time at which the image was received by the mail server, and the completion time gives the time at which the images were uploaded on the web-site and hence available to the user.}}
\label{imagetable}
\end{center}
\end{table}
\end{spacing}

\end{document}